\documentclass[letterpaper, 10 pt, journal, twoside]{IEEEtran}
\usepackage{amsmath,amsfonts}
\usepackage{algorithmic}
\usepackage{algorithm}
\usepackage{array}
\usepackage[caption=false,font=normalsize,labelfont=sf,textfont=sf]{subfig}

\usepackage{textcomp}
\usepackage{stfloats}
\usepackage{url}
\usepackage{hyperref}
\usepackage{verbatim}
\usepackage{graphicx}
\usepackage{cite}
\usepackage{multirow}
\usepackage{amssymb}
\usepackage{booktabs}
\usepackage{arydshln}
\usepackage{needspace}
\usepackage{marvosym}
\usepackage{hhline}
\usepackage{colortbl}
\usepackage{tabularx}
\usepackage{xcolor}
\newcommand{\zhiyuanrevision}[1]{{\color{black}#1}}

\begin{document}

\title{VTLoc: Learning-based Tactile Contact Localization in Visual Point Clouds}

\author{
Zhiyuan Wu, Zhuo Chen, and Shan Luo
\thanks{This work was supported in part by the EPSRC project ``TacDiff: Designing Tactile-based Robots via Differentiable Simulations'' (UKRI3438). (Corresponding author: Shan Luo.)}
\thanks{Zhiyuan Wu, Zhuo Chen, and Shan Luo are with Department of Engineering, King's College London, Strand, London, WC2R 2LS, United Kingdom, \{zhiyuan.1.wu, zhuo.7.chen, shan.luo\}@kcl.ac.uk. }
}


\maketitle


\begin{abstract}
Vision and touch are complementary modalities essential for robotic perception and manipulation. While vision provides global object context, touch offers precise local information at contact points. Integrating these modalities for contact localization, i.e., predicting the location of touch on an object's surface, poses significant challenges due to the need for accurate spatial alignment between tactile data and visual geometry. To address this challenge, we propose VTLoc, a novel visual-tactile framework that localizes contact points from tactile readings using a 3D point cloud as visual input. VTLoc introduces two key components: a geometric multi-modal alignment module, which reconstructs a pseudo-point cloud from fused visual-tactile features and aligns it with the visual point cloud to enforce spatial consistencies across modalities; and an iterative localizing updater, which iteratively refines the predicted contact location using fused visual-tactile features. Evaluated on a new benchmark of 100 real-world objects, VTLoc improves single-touch contact localization by reducing local-to-global correspondence ambiguity. Code and data are available on: \href{https://georgewuzy.github.io/vtloc-website/}{https://georgewuzy.github.io/vtloc-website/}. 
\end{abstract}



\section{Introduction} 
\label{sec:introduction} 
Vision and touch are two fundamental sensory modalities for robots, offering complementary information that enhances perception and manipulation capabilities \cite{wu2025vitacgen, wu2025cedex}. Vision provides a global understanding of objects and their environments by capturing shape, texture, and spatial relationships, while touch delivers fine-grained local information about surface geometry and contact forces \cite{chen2026genforce}. Recent  research has increasingly focused on fusing visual and tactile data to leverage these complementary strengths \cite{luo2018vitac, wu2025convitac}. However, many existing approaches rely primarily on direct image or feature concatenation \cite{luo2017tactilesurvey3}, without explicitly addressing the challenge of localizing tactile perception within the visual scene.

Given an object's visual representations like a mesh or point cloud, the location on its surface where contact occurs can be predicted from tactile readings \cite{gao2023objectfolderreal}. This task, known as \textit{contact localization}, has attracted growing attention in recent years \cite{corcoran2010measurement, molchanov2016contact},  as it is crucial for accurate object recognition, manipulation, and interaction. Single-touch contact localization provides a contact estimate from a single tactile observation that can be used to initialize and refine localization in multi-touch and sequential tactile localization frameworks\cite{suresh2023ycbslide}.

Existing contact localization methods can be divided into same-dimensional and cross-dimensional matching. In same-dimensional approaches, the input and output spaces share the same dimensionality, e.g., mapping 2D tactile images to 2D contact points in visual images \cite{luo2015localizing, villalonga2021tactile} or aligning 3D tactile point clouds with 3D object surfaces \cite{li2024hypertaxel}. In contrast, cross-dimensional matching involves localizing 2D tactile images within a 3D scene \cite{suresh2023ycbslide, gao2023objectfolderreal}, which is more challenging as it requires learning  mappings across different spatial dimensions. Prior work in cross-dimensional matching often relies on strong data priors or constraints, \textit{e.g.}, constructing a large tactile codebook from randomized sensor poses sampled across an object mesh \cite{suresh2023ycbslide}, which limits applicability in real-world settings due to high data and computation requirements. 

To address the challenges of cross-modal contact localization, ObjectFolder Real \cite{gao2023objectfolderreal} introduces a benchmark of localizing 2D tactile images on the surface of 3D point clouds. 
Building on this foundation, we propose VTLoc (\underline{\textbf{V}}isual \underline{\textbf{T}}actile  \underline{\textbf{Loc}}alization), a novel framework that predicts contact coordinates on an object's surface from \underline{\textbf{tactile}} readings, using the object's 3D point cloud as \underline{\textbf{visual}} input, which can be easily obtained via an RGB-D camera or 3D scanner. 
Inspired by human perception, where an internal object representation is formed and refined through sensory feedback\cite{lange2024tactilememory}, VTLoc introduces two key modules: The geometric multi-modal alignment (GMA) module reconstructs a pseudo-point cloud from visual and tactile features and aligns it with the original point cloud, to enhance the geometric consistency and spatial reasoning of both modalities; The iterative localizing updater (ILU) performs iterative contact location regression, refining the prediction by matching tactile features with visual input. The final contact location is selected from a candidate contact set, and a 3D contact probability heat-map is generated to indicate the likelihood of contact at each point in the point cloud. To validate our proposed VTLoc method, we construct a new contact localization benchmark based on ObjectFolder Real \cite{gao2023objectfolderreal}, comprising 100 real-world objects. Qualitative and quantitative results on this benchmark demonstrate the effectiveness of the proposed VTLoc. 

\begin{figure*} [t!]
	\centering
    \vspace{0.8em} 
	\includegraphics[width=0.6 \textwidth]{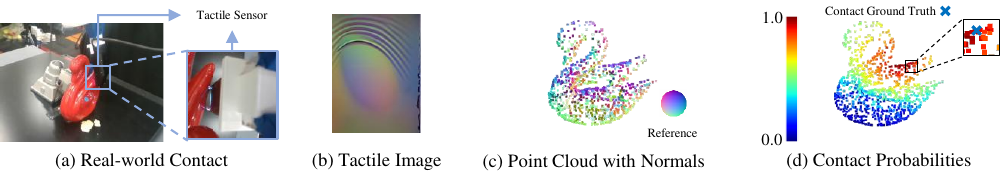}
    \vspace{-0.8em} 
	\caption{
        An example of a swan in the ObjectFolder Real \cite{gao2023objectfolderreal} dataset, demonstrating the effectiveness of our VTLoc method. (a) Tactile contact in the real world. (b) Tactile image collected from the tactile sensor during interaction with the object. (c) Point cloud with normals. (d) Predicted probability heat-map and the corresponding contact ground truth for reference. 
    }\label{fig:intro}
    \vspace{-1.2em} 
\end{figure*}

The key contributions of this work are as follows:
\begin{itemize}
    \item We propose VTLoc, a novel framework for visual-tactile contact localization that predicts surface coordinates from tactile readings, using a 3D point cloud as visual input. 
    \item We design two core modules: a geometric multi-modal alignment module that enhances spatial consistency between tactile and visual modalities, and an iterative localizing updater that iteratively refines contact predictions. 
    \item We establish a new benchmark for contact localization with 100 real-world objects and demonstrate the effectiveness of VTLoc through extensive experiments.
\end{itemize}

\section{Related Works} 
\label{sec:related_works}
Contact localization has been a long-standing challenge in robotic perception and multi-modal sensing \cite{gao2023objectfolderreal}. Early efforts, such as Corcoran \textit{et al.} \cite{corcoran2010measurement} and Petrovskaya \textit{et al.} \cite{petrovskaya2011global}, introduced probabilistic Bayesian frameworks to estimate contacts over an object's surface. Li et al. \cite{li2014localization} proposed to localize tactile images with a 
height map via image registration to help localize objects in 
hand. Furthermore, Bimbo \textit{et al.} \cite{bimbo2016hand}, for instance, used the covariance of tactile sensor arrays to infer contact pose of objects. Recent efforts have focused on tactile representation learning for more accurate and generalizable contact localization. These methods can be classified into two categories: 

1) \textbf{Same-dimensional matching}, where the dimensionality of the tactile input matches the output space. For example, Luo \textit{et al.} \cite{luo2015localizing} employed recursive Bayesian filtering to localize a 2D tactile image within a 2D visual map. Subsequent works such as those by Bauza \textit{et al.} \cite{villalonga2021tactile, bauza2023tac2pose} leveraged geometric contact rendering to refine 2D contact localization. Bauza \textit{et al.} \cite{bauza2019imprints} localizes contact objects through Iterative Closest Point (ICP)-based pose estimation and tactile imprint matching. More recently, Li \textit{et al.} \cite{li2024hypertaxel} proposed a contrastive learning framework to embed 3D tactile signals and object surfaces in a shared feature space. 

2) \textbf{Cross-dimensional matching}, where the dimensionality of the tactile input differs from the output space. This scenario is especially relevant for vision-based tactile sensors \cite{yuan2017gelsight}, which produce 2D tactile images but require contact localization in 3D space. For example, MidasTouch \cite{suresh2023ycbslide} performs localization by querying a precomputed tactile codebook generated from 50,000 randomized sensor poses on each object mesh. While effective, such methods rely heavily on object-specific priors and offline sampling, limiting scalability and real-world applicability. Moreover, they only have 10 objects with merely five trajectories collected per object. More recently, GelSLAM \cite{huang2025gelslam} developed by Huang \textit{et al.} performs tactile-based contact localization on objects through spatial alignment of tactile readings using differential representations.

\begin{table*}[!t]
    \centering\footnotesize{
    \vspace{0.8em} 
	\caption{
            Comparison of various visual-tactile contact localization works. 
    }
    \vspace{-0.8em} 
\label{tab:related_works}
    \begin{tabular}{l|c|c}
    \toprule
        Works & Input & Output \\
        \hline
        \cite{corcoran2010measurement, molchanov2016contact} & Sensor data \& Sensor configuration & Contact location in robot hand \\
        \cite{bimbo2016hand} & Sensor data \& Sensor pose \& Object point cloud & Contact pose of object \\
        \cite{luo2015localizing} & Tactile image \& visual image & Contact location in visual image \\
        \cite{villalonga2021tactile, bauza2023tac2pose, li2024hypertaxel} & Tactile image \& Pre-computed poses & Contact pose distribution \\
        \cite{suresh2023ycbslide} & Tactile image \& Pre-computed tactile codebook & Contact location on object \\
        \cdashline{1-3}
        \cite{gao2023objectfolderreal}, \textbf{Ours} & \textbf{Tactile image} \& \textbf{Object point cloud} & \textbf{Contact location on object} \\
    \bottomrule
    \end{tabular}
    \vspace{-0.6em} 
}\end{table*}

In contrast to prior work, as summarized in Table \ref{tab:related_works}, our proposed VTLoc framework addresses these limitations by directly learning to predict contact locations from 2D tactile images and 3D point clouds without reliance on dense object-specific sampling. By introducing a Geometric Multi-Modal Alignment module and an Iterative Localizing Updater, VTLoc explicitly models the spatial relationship between tactile and visual data and progressively refines contact predictions, achieving robust localization across diverse object geometries.

\section{Learning-based Tactile Contact Localization in Visual Point Cloud}  
\label{sec:methodology}  

\subsection{Problem Setting}
\label{sec:problem_setting}
\begin{figure} [!t]
	\centering
	\includegraphics[width=0.5 \linewidth]{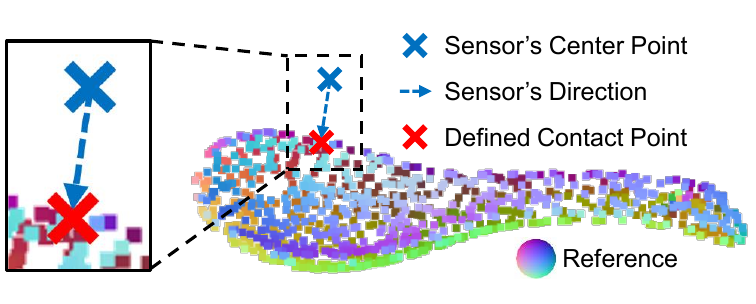}
    \vspace{-0.8em}
	\caption{
        The contact point is defined as the projected point of the tactile sensor's center point along its surface normal onto the object. 
        }
    \label{fig:projection}
    \vspace{-0.8em}
\end{figure}
We refer to the same problem setting of contact localization in ObjectFolder \cite{gao2023objectfolderreal}: \textit{Given the object's mesh and sensory observations of the contact position, contact localization aims to predict the vertex coordinate of the surface location on the mesh where the contact happens.} 
Specifically, we define the “contact point” as the projected point of the center of the tactile sensor surface on the mesh of the object along the surface normal, as shown in Fig. \ref{fig:projection}. 
We consider a robot equipped with a  vision-based tactile sensor that captures RGB tactile images from real-world object interactions. As shown in Fig. \ref{fig:intro}, the system receives two inputs:

1) A tactile image $\boldsymbol{T} \in \mathbb{R}^{H \times W \times 3}$, where $H$ and $W$ represent the height and width of the tactile image. Each pixel in $\boldsymbol{T}$ represents a specific location on the tactile sensor. 

2) A 3D point cloud $\boldsymbol{P} \subset \mathbb{R}^{3}$ representing the target object, containing $N^p$ points.
The point cloud can be complete or partial, easily calculated via an RGB-D camera or 3D scanner. Each point $\boldsymbol{p} \in \boldsymbol{P}$ is defined as:  
\begin{equation}
    \boldsymbol{p} = (x, y, z), 
\end{equation}
where $(x, y, z)$ are the spatial coordinates of the point in 3D space. In addition, each point is associated with a surface normal vector $(n_x, n_y, n_z)$. 

Based on retrieval frameworks in \cite{gao2023objectfolderreal, wi2023virdo++}, we define a contact set $\boldsymbol{S}^c \subset \mathbb{R}^{3}$, which contains all ground truth contact points where contact data were collected around each object:
\begin{equation}
    \boldsymbol{S}^c = \{ \boldsymbol{c}^{gt} = (x_c, y_c, z_c) \mid x_c, y_c, z_c \in \mathbb{R} \},
\end{equation}
where $\boldsymbol{c}$ refers to any ground truth contact point. The objective is to predict the contact coordinates $\boldsymbol{c}^p$ on the object's surface where the contact occurs and a corresponding 3D contact probability map $\boldsymbol{P}^s \subset \mathbb{R}^{3}$:
\begin{equation}  
    \boldsymbol{P}^s = \{ (\boldsymbol{p}, s_{\boldsymbol{p}}) \mid \boldsymbol{p} \in \boldsymbol{P} \}, 
\end{equation}  
where $s_{\boldsymbol{p}}$ denotes the predicted likelihood that point $\boldsymbol{p}$ is the actual contact location.

\subsection{Network Overview}
\begin{figure*} [t!]
	\centering
	\includegraphics[width=0.75 \textwidth]{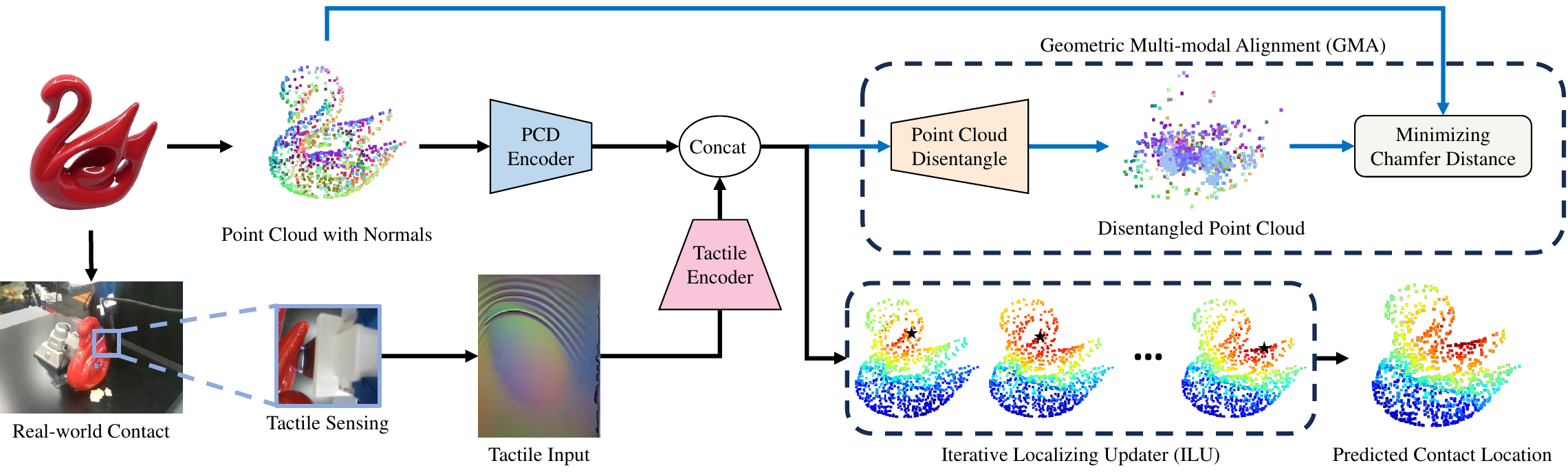}
    \vspace{-0.8em} 
	\caption{
        VTLoc workflow that predicts coordinates on an object's surface from tactile readings where the contact occurs. The framework includes a geometric multi-modal alignment (GMA) module, which enhances geometric consistency by minimizing the Chamfer distance between the pseudo-point cloud reconstructed from fused visual and tactile features and the original point cloud. An iterative localizing updater is introduced to perform iterative refinement to predict the contact prediction $\boldsymbol{c}^p$ (denoted by a star). This prediction is used to retrieve the top-K most likely contact candidates  $\{ \boldsymbol{c}^*_1, \dots, \boldsymbol{c}^*_k \}$ from the contact set $\boldsymbol{S}^c$. Finally, a 3D contact probability heat-map is generated over the point cloud, reflecting the likelihood of contact at each point. 
    }\label{fig:pipeline}
    \vspace{-1.2em} 
\end{figure*}

As illustrated in Fig. \ref{fig:pipeline}, our model takes as input a tactile contact image $\boldsymbol{T}$ and a point cloud $\boldsymbol{P}$ that represents the object geometry and contains coordinates and surface normals. To extract meaningful representations, we employ a ResNet-18 \cite{he2016resnet} followed by a flattening Multi-Layer Perceptron (MLP) to encode the tactile input, producing a tactile feature vector $\boldsymbol{F}^t \in \mathbb{R}^C$, where $C$ denotes the number of feature channels. Simultaneously, we utilize PointNet++ \cite{qi2017pointnet++} as a point cloud encoder to a corresponding feature vector $\boldsymbol{F}^p \in \mathbb{R}^C$. 

The extracted tactile and visual representations are then fused and passed through two key modules: the geometric multi-modal alignment (GMA) module and the iterative localization updater (ILU) module. The GMA module aligns the fused features in 3D space to enforce geometric consistency across modalities, while the ILU performs iterative regression to refine the predicted contact location $\boldsymbol{c}^p$. 

\subsection{Geometric Multi-Modal Alignment}  
Human contact localization follows a perceptual paradigm: mentally reconstructing an object, then iteratively matching the touched region using tactile memory \cite{lange2024tactilememory}. Inspired by this paradigm, we first introduce a \textbf{G}eometric \textbf{M}ulti-modal \textbf{A}lignment (GMA) module to reconstruct a pseudo point cloud from visual and tactile features to align visual and tactile modalities. 

The GMA module takes the fused feature vector $\boldsymbol{F}^f = [\boldsymbol{F}^t, \boldsymbol{F}^p]$ as input, and reconstructs a pseudo-point cloud $\boldsymbol{P}^d$ from this representation. This reconstruction serves to preserve and reintroduce explicit geometric structure from the fused visual-tactile features. The reconstructed $\boldsymbol{P}^d$ is then aligned with the input point cloud $\boldsymbol{P}$ by minimizing the Chamfer Distance, encouraging the tactile and visual encoders to learn geometrically consistent representations. 

This disentangling process maps $\boldsymbol{F}^f$ to a structured point cloud representation through learned weights $\boldsymbol{W}^d \in \mathbb{R}^{N^d \times 2C \times C^d}$: 
\begin{equation}  
\boldsymbol{F}^d(j,l) = \sum \boldsymbol{F}^f(i) \cdot \boldsymbol{W}^d(i,j,l),  
\end{equation}  
where $\boldsymbol{F}^d$ is the disentangled pseudo-point cloud, and $N^d$ and $C^d$ denote the number of reconstructed points and the number of output channels, respectively.

An MLP then decodes each feature vector in $\boldsymbol{F}^d$ into a 3D point with an associated normal, yielding $\boldsymbol{p}^d = [x^d, y^d, z^d, n_x^d, n_y^d, n_z^d]$. We optimize a reconstruction loss between $\boldsymbol{P}^d$ and $\boldsymbol{P}$ (see Sec. \ref{method:loss} for details) to enforce spatial alignment between the tactile and point-cloud modalities in both spatial coordinates and surface normals. This process enhances the 3D geometric understanding of the fused tactile and visual modalities.  

\subsection{Iterative Localization Updater}  
Humans refine contact localization through an iterative process, comparing current sensory feedback against stored tactile memory to gradually improve precision \cite{lange2024tactilememory}. 
Inspired by this mechanism, we propose the \textbf{I}terative \textbf{L}ocalizing \textbf{U}pdater (ILU), a module that mimics this iterative refinement by generating a sequence of contact predictions during training. 

As illustrated in Fig.~\ref{fig:pipeline}, the ILU begins from an initial estimate $\hat{\boldsymbol{c}}_0 = \boldsymbol{0}$, and predicts a sequence of contact locations $\{ \hat{\boldsymbol{c}}_1, \dots, \hat{\boldsymbol{c}}_N \}$ over $N$ iterations. At each step $k$, the model refines its estimate by computing an update direction $\Delta \hat{\boldsymbol{c}}$, applied as:
\begin{equation}  
\hat{\boldsymbol{c}}_{k+1} = \hat{\boldsymbol{c}}_k + \Delta\hat{\boldsymbol{c}}.  
\end{equation}  

The ILU takes the tactile feature $\boldsymbol{F}^t$, the point cloud feature $\boldsymbol{F}^p$, and a hidden state $\boldsymbol{h}_k$ as inputs, producing both an updated $\Delta\hat{\boldsymbol{c}}$ and an updated hidden state $\boldsymbol{h}_{k+1}$. Drawing inspiration from iterative refinement strategies in matching tasks, such as optical flow estimation and stereo matching \cite{wu2024s3mnet, wu2024sgroadseg}, we implement the ILU using a linear Gated Recurrent Unit (GRU). 

The initial hidden state is initialized as $\boldsymbol{h}_0 = \mathcal{T}(\boldsymbol{F}^t)$, where $\mathcal{T}$ represents a $tanh$ activation that maps the feature values to the range $[-1, 1]$. At each iteration, the hidden state is updated via:  
\begin{equation}  
\boldsymbol{h}_{k+1} = (\boldsymbol{1} - \boldsymbol{z}_k) \odot \boldsymbol{h}_k + \boldsymbol{z}_k \odot \boldsymbol{q}_k,  
\end{equation}  
where $\odot$ denotes element-wise multiplication, and $\boldsymbol{z}_k$ is the update gate, defined as:  
\begin{equation}  
\boldsymbol{z}_k = \sigma \left( \text{Linear}^z \left( [\boldsymbol{h}_k, \boldsymbol{x}] \right) \right),  
\end{equation}  
with   
\begin{equation}  
\boldsymbol{x} = [\boldsymbol{F}^p, \text{ReLU}(\boldsymbol{F}^t)].  
\end{equation}

The candidate hidden state $\boldsymbol{q}_k$ is computed as:  
\begin{equation}  
\boldsymbol{q}_k = \mathcal{T} \left( \text{Linear}^q \left( [\boldsymbol{r}_k \odot \boldsymbol{h}_k, \boldsymbol{x}] \right) \right),  
\end{equation}  
where $\boldsymbol{r}_k$ represents the reset gate, expressed as:  
\begin{equation}  
\boldsymbol{r}_k = \sigma \left( \text{Linear}^r \left( [\boldsymbol{h}_k, \boldsymbol{x}] \right) \right).  
\end{equation}  

For each hidden state $\boldsymbol{h}_k$, an MLP layer serves as a decoder to derive $\Delta\hat{\boldsymbol{c}}$ for the contact location update. During inference, the final predicted location $\hat{\boldsymbol{c}}_N$ is utilized as the predicted contact location $\boldsymbol{c}^p$ for matching with $\boldsymbol{S}^c$. The cumulative loss is computed as the sum of prediction errors at all intermediate steps, enabling the ILU to learn accurate and stable refinement trajectories (see Sec.~\ref{method:loss} for details). At inference time, the final estimate $\hat{\boldsymbol{c}}_N$ is used as the predicted contact location $\boldsymbol{c}^p$ for matching within the candidate contact set $\boldsymbol{S}^c$.

\subsection{Contact Probability Calculation}
To get the 3D probability map from the predicted contact position $\boldsymbol{c}^p$, we calculate a contact score $S_{\boldsymbol{c}^*_i}$ for every potential contact points $\boldsymbol{c}^*_i$ in the contact set $\boldsymbol{S}^c$ as: 
\begin{equation}  \label{eq:distance}
    S_{\boldsymbol{c}^*_i} = \left\| \boldsymbol{c}^{p} - \boldsymbol{c}^*_i \right\|^2,  
\end{equation}  
where $\left\| \cdot \right\|$ denotes the Euclidean distance that quantifies the discrepancy between $\boldsymbol{c}^{p}$ and the points in $\boldsymbol{S}^c$. We can therefore obtain the Top-K optimal contact locations $\boldsymbol{C}^* = \{ \boldsymbol{c}^*_1, \dots, \boldsymbol{c}^*_K \}$ by ranking $S_{\boldsymbol{c}^*_i}$ in ascending order.

For each point $\boldsymbol{p} \in \boldsymbol{P}$, we find the nearest contact candidate $\boldsymbol{c}^*_{\boldsymbol{p}}$ from $\boldsymbol{C}^*$ as:
\begin{equation}    
\boldsymbol{c}^*_{\boldsymbol{p}} = \arg \min_{\boldsymbol{c}^*_i \in \boldsymbol{C}^*} \left\| \boldsymbol{p} - \boldsymbol{c}^*_i \right\|^2,
\end{equation}  
and we have a weighted index $w$ as: 
\begin{equation}
    w = \arg \min_{j} \left\| \boldsymbol{p} - \boldsymbol{c}^*_j \right\|^2.
\end{equation}
We can compute a contact score $s_{\boldsymbol{p}}$ for each point $\boldsymbol{p}$ by calculating a weighted distance:
\begin{equation}
    s_{\boldsymbol{p}} = \frac{1}{w} e^{-\left\| \boldsymbol{p} - \boldsymbol{c}^*_w \right\|} ,
\end{equation}
Therefore, we have a 3D probability heat-map $\boldsymbol{P}^s \subset \mathbb{R}^{3}$:
\begin{equation}  
    \boldsymbol{P}^s = \{ (\boldsymbol{p}, s_{\boldsymbol{p}}) \mid \boldsymbol{p} \in \boldsymbol{P} \}, 
\end{equation}  
and further a normalized heat-map $\hat{\boldsymbol{P}}$ within $0$ to $1$:
\begin{equation}
\label{eq:norm}
     \hat{\boldsymbol{P}} = \frac{\boldsymbol{P}^s - min(\boldsymbol{P}^s)}{max(\boldsymbol{P}^s) - min(\boldsymbol{P}^s)}. 
\end{equation}
In $\hat{\boldsymbol{P}}$, each point is assigned a probability score that reflects its likelihood of being a contact location. 

\subsection{Loss Function} \label{method:loss}
The loss function for our model comprises a reconstruction loss $\mathcal{L}^{recon}$ for the GMA module, calculated using Chamfer Distance (CD) \cite{fan2017cd}, expressed as:  
\begin{equation}  
\mathcal{L}^{recon} = \sum_{\boldsymbol{p}^d \in \boldsymbol{P}^d} \min_{\boldsymbol{p} \in \boldsymbol{P}} \|\boldsymbol{p}^d - \boldsymbol{p}\|^2 + \sum_{\boldsymbol{p} \in \boldsymbol{P}} \min_{\boldsymbol{p}^d \in \boldsymbol{P}^d} \|\boldsymbol{p} - \boldsymbol{p}^d\|^2.  
\end{equation}  
Additionally, we incorporate a sequence loss $\mathcal{L}^{seq}$ for the ILU, formulated as:  
\begin{equation}  
\mathcal{L}^{seq} = \sum_{i=1}^{N} \gamma^{N-i} \left\|\boldsymbol{c}^{gt}-\boldsymbol{c}_{i}\right\|,  
\end{equation}  
where $\boldsymbol{c}^{gt}$ represents the ground-truth contact location, and $\gamma^{N-i}$ denotes exponentially increasing weights, with $\gamma$ set to $0.9$. Consequently, the overall loss function is expressed as:  
\begin{equation}  
\mathcal{L} = \mathcal{L}^{seq} + \lambda \mathcal{L}^{recon},
\end{equation}
where $\lambda$ is empirically set to $1$.

\section{Experiments}
\label{sec:experiments}

\subsection{Dataset}
\begin{figure*} [!t]
	\centering
    \vspace{0.8em} 
	\includegraphics[width=0.8 \textwidth]{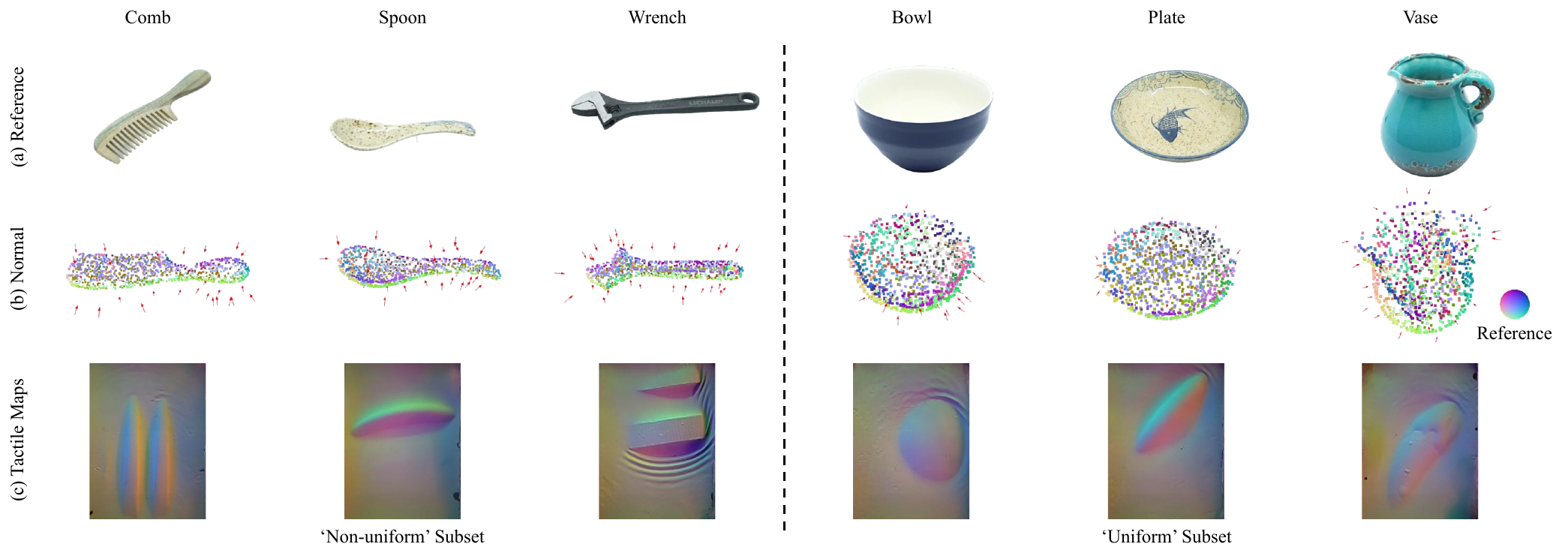}
    \vspace{-1.2em} 
    \caption{
        Examples of resampled data in our proposed benchmark. We choose 100 daily objects from ObjectFolder Real \cite{gao2023objectfolderreal} and resample them to get point clouds of both positions and normals, with 30-50 contact locations for each object. The dataset is quantitatively divided into a `non-uniform' subset, whose objects are characterized by distinct surface curvature variations, and a `uniform' subset, whose objects exhibit more uniform surface curvature variations. 
    }\label{fig:data}
    \vspace{-1.2em} 
\end{figure*}

We develop our benchmark based on the ObjectFolder Real \cite{gao2023objectfolderreal}. As shown in examples in Fig. \ref{fig:data}, we select 100 real-world daily objects and resample their point clouds of both positions and normals, with 30-50 contact locations for each object. We split our data into training and testing sets at the data level using a 7:3 ratio. From the test set, we select 9-15 representative points per object as discrete contact candidates. Referring to \cite{gao2023objectfolderreal}, for each object, an EinScan Pro HD 2020 handheld 3D scanner is utilized to capture high-quality 3D meshes and corresponding color textures in a controlled dark environment. The scanner achieves high accuracy by projecting a visible light array onto the object and recording the texture through an integrated camera. To effectively obtain point cloud data, we randomly sample 1,024 points from each 3D mesh, capturing their spatial coordinates and normal vectors, as illustrated in Fig. \ref{fig:data} (b). The tactile data are collected using a Franka Emika Panda robot arm equipped with a GelSight tactile sensor \cite{yuan2017gelsight}, which features a sensing area of $32 \times 24 \ mm^2$. For each object, suitable surrounding contact points are selected. 
In the tactile data collection setup, the robot arm employs position-controlled motion along the surface normal of each target point. To ensure both hardware safety and data reproducibility, the motion stops at a predefined operational threshold, \textit{e.g.}, a safe penetration depth of $\sim$1.0 mm into the $\sim$2 mm thick silicone layer \cite{yuan2017gelsight}. This criterion maximizes the compression for high-fidelity geometry capture while strictly avoiding the underlying acrylic plate to prevent irreversible damage. Four RealSense cameras (with standard calibration) were used for alignment between contact points and the object mesh using the Iterative Closest Point (ICP) algorithm. 
 
To avoid contact ambiguity of multiple possible contact points that can produce similar tactile readings, we divide the objects into ``non-uniform'' and ``uniform'' subsets according to their normal variation. For each normalized point cloud $P$ of an object, we compute the local normal variation as:
\begin{equation}
    \Delta n = \frac{||{n}_{i+1} - {n}_i||}{||{p}_{i+1} - {p}_i||}, 
\end{equation}
where ${n}_i$ and ${p}_i$ are the normal vector and position at vertex $i$. 16 objects are classified as ``non-uniform'', where less than 70\% of their points are below a normal variation threshold of 30, indicating a lower degree of smoothness for the surface and less ambiguity, \textit{e.g.}, comb, spoon, fork, and wrench. The remaining 84 objects are classified as ``uniform''. 

\subsection{Evaluation Metrics}
We first evaluate the millimeter error (mm Error) and the normalized distance (ND) between the predicted contact position $\boldsymbol{c}^{p}$ and the ground truth contact position $\boldsymbol{c}^{gt}$, which can be defined as $\text{mm Error} = \|\boldsymbol{c}^{p} - \boldsymbol{c}^{gt}\|$ and $\text{ND} = \frac{\|\boldsymbol{c}^{p} - \boldsymbol{c}^{gt}\|}{D}$. $D$ is the object scale, \textit{i.e.}, the maximum distance between any two points of the object point cloud. ND provides a scale-invariant measure across objects of varying dimensions, \textit{e.g.}, a small pot and a large pan.

Second, we evaluate Top-1 and Top-5 accuracy (Top-1 Acc, Top-5 Acc). The Top-1 Acc measures whether the predicted contact position $\boldsymbol{c}^p$ is closest to the ground truth contact candidate among the discrete set of test candidates described in Sec. IV.A. Top-5 Acc is defined similarly for the five nearest candidates. 

Finally, for the normalized 3D probability heat-map $\hat{\boldsymbol{P}}$, we evaluate its mean probability error (MPE) against the normalized ground truth probability heat-map $\hat{\boldsymbol{P}}^{gt} \subset \mathbb{R}^3$, whose every point score $s_{\boldsymbol{p}}^{gt}$ is obtained as:
\begin{equation}
    s_{\boldsymbol{p}}^{gt} = e^{-\left\| \boldsymbol{p} - \boldsymbol{c}^{gt} \right\|}.
\end{equation}
The MPE is calculated as the mean absolute error between the $\hat{\boldsymbol{P}}$ and $\hat{\boldsymbol{P}}^{gt}$ across all points. 

\begin{figure*} [t!]
	\centering
    \vspace{0.8em} 
	\includegraphics[width=0.58 \textwidth]{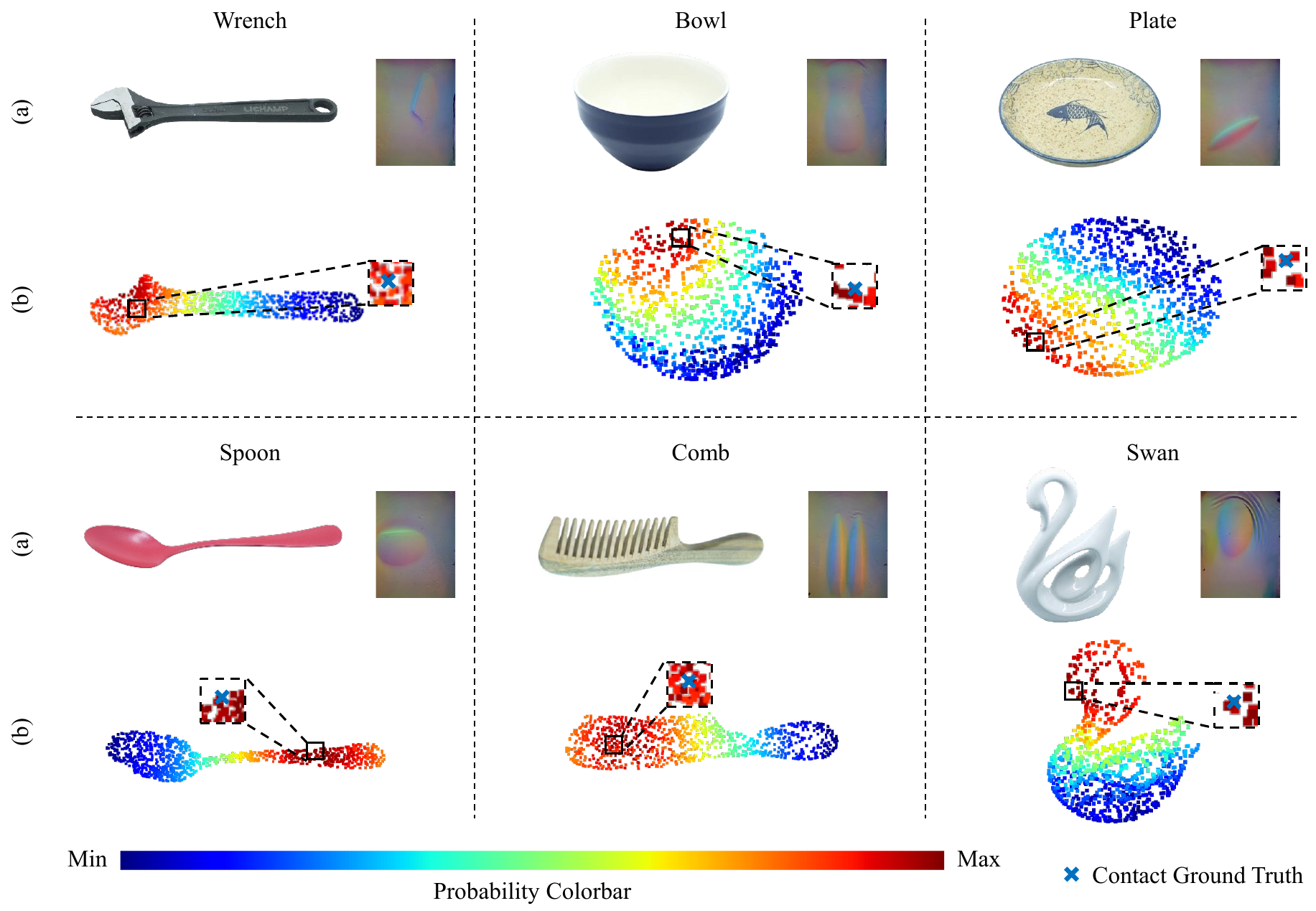}
    \vspace{-1.0em} 
	\caption{
        Qualitative results of contact probabilities. (a) Reference and contact tactile images of objects in real world. (b) 3D probability heat-map with highlighted contact ground truth. VTLoc exhibits robust performance in connecting visual and tactile modalities, even in challenging cases such as the intricate geometry of the swan. More results are available in the demo video. 
    }\label{fig:viz_prob}
    \vspace{-0.8em} 
\end{figure*}

\subsection{Comparison with Baselines} \label{exp:comparison}  
We compare our VTLoc with several baselines for visual-tactile contact localization, including the Point Filtering method and the  Multi-sensory Contact Regression (MCR) method from the ObjectFolder benchmark \cite{gao2023objectfolderreal}, and MidasTouch \cite{suresh2023ycbslide} in the single-contact setup.
We first compared them in our `non-uniform' subset. For objects with non-uniform surface curvature variations, each tactile image corresponds to a unique location on object without ambiguity, directly reflecting the performance of tactile contact localization algorithms. As reported in Tab. \ref{tab:comparison}, when only coordinates are used as point cloud input, VTLoc outperforms all baselines. In particular, VTLoc improves ND by 14.87\% compared to the Point Filtering method and by 3.11\% compared to MCR that achieves the best among baselines due to its fusion of point cloud and tactile features. For Top-5 Acc, VTLoc improves by 30.19\% compared to the Point Filtering method, and by 5.66\% compared to MCR.

These gains stem from our proposed ILU, which effectively decouples the two modalities by converging updates toward a consistent predicted point, along with a GMA module that enhances the network’s geometric understanding. Furthermore, incorporating normals into the point cloud (in addition to coordinates) yields further improvements. Specifically, compared to taking only coordinates as input, VTLoc with both coordinates and normals further improves ND by 0.74\%, Top-1 Acc by 6.91\%, Top-5 Acc by 1.88\%, and MPE by 7.16\%, highlighting the importance of richer geometric input for tactile contact localization.

Next, we evaluated the performance of VTLoc on the `uniform' subset. As shown in Tab. \ref{tab:comparison_uniform}, VTLoc achieves significant improvements on objects with uniform surface curvature variations. In particular, VTLoc reduces ND by 12.11\% compared to the Point Filtering method, and by 10.01\% compared to MCR. For Top-5 Acc, VTLoc improves by 25.35\% compared to the Point Filtering method and by 9.16\% compared to \zhiyuanrevision{MidasTouch} that performs the best among baselines. However, VTLoc's performance on this subset remains relatively lower compared to the `non-uniform' subset, due to the inherent geometric ambiguity of uniform objects. Interestingly, adding normals as input further degrades VTLoc's performance on the `uniform' subset. This is likely because the geometric information of uniform objects does not offer meaningful cues for tactile-based localization, making additional normals either ineffective or even counterproductive in this scenario.

\begin{table*}[t!]
	\centering\footnotesize{
	\caption{
        Quantitative results of our proposed VTLoc compared with other baselines in our `non-uniform' subset. 
    }
    \vspace{-0.8em} 
\label{tab:comparison}
	\begin{tabular}
            {l c c c c c c c c c c}
            \toprule
            \multicolumn{1}{c}{\multirow{2}{*}{Methods}} & \multicolumn{2}{c}{Modality} & {\multirow{2}{*}{mm Error (mm) $\downarrow$}} & \multicolumn{1}{c}{\multirow{2}{*}{ND (\%) $\downarrow$}} & \multicolumn{1}{c}{\multirow{2}{*}{Top-1 Acc ($\%$) $\uparrow$}} & \multicolumn{1}{c}{\multirow{2}{*}{Top-5 Acc ($\%$) $\uparrow$}} & \multicolumn{1}{c}{\multirow{2}{*}{MPE $\downarrow$}} \\
            & Coordinates & Normals \\
            \hline
            Chance &-&-& - & - & 12.58 & 50.94 & 0.2940 \\
            \cdashline{1-8}
            \multicolumn{1}{l}{\multirow{1}{*}{Point Filtering}} & \checkmark &  & 71.81 & 35.78 & 11.32 & 54.09 & 0.2788 \\
            \multicolumn{1}{l}{\multirow{1}{*}{MCR}} & \checkmark &  & 46.82 & 24.02 & 27.04 & 78.62 & 0.1441 \\
            MidasTouch & - & - & 50.01 & 26.74 & 36.48 & 76.73 & 0.1594 \\
            \cdashline{1-8}
            Ours & \checkmark &  & 39.24 & 20.91 & 37.74 & 84.28 & 0.1117 \\
            Ours & \checkmark & \checkmark & \textbf{37.57} & \textbf{20.17} & \textbf{44.65} & \textbf{86.16} & \textbf{0.1037} \\
    	\bottomrule
		\end{tabular}
    \vspace{-1.2em} 
}\end{table*}

\begin{table*}[t!]
	\centering\footnotesize{
	\caption{
        Quantitative results of our proposed VTLoc compared with other baselines in our `uniform' subset. 
    }
    \vspace{-0.8em} 
\label{tab:comparison_uniform}
	\begin{tabular}
            {l c c c c c c c c c c}
            \toprule
            \multicolumn{1}{c}{\multirow{2}{*}{Methods}} & \multicolumn{2}{c}{Modality} & {\multirow{2}{*}{mm Error (mm) $\downarrow$}} & \multicolumn{1}{c}{\multirow{2}{*}{ND (\%) $\downarrow$}} & \multicolumn{1}{c}{\multirow{2}{*}{Top-1 Acc ($\%$) $\uparrow$}} & \multicolumn{1}{c}{\multirow{2}{*}{Top-5 Acc ($\%$) $\uparrow$}} & \multicolumn{1}{c}{\multirow{2}{*}{MPE $\downarrow$}} \\
            & Coordinates & Normals \\
            \hline
            Chance &-&-& - & - & 6.26 & 36.54 & 0.2599 \\
            \cdashline{1-8}
            \multicolumn{1}{l}{\multirow{1}{*}{Point Filtering}} & \checkmark &  & 95.31 & 48.82 & 7.62 & 39.01 & 0.2404 \\
            \multicolumn{1}{l}{\multirow{1}{*}{MCR}} & \checkmark &  & 87.47 & 46.72 & 16.31 & 51.06 & 0.2039 \\
            MidasTouch & - & - & 88.56 & 47.26 & 18.79 & 55.20 & 0.2114 \\
            \cdashline{1-8}
            Ours & \checkmark & & \textbf{68.40} & \textbf{36.71} & \textbf{21.79} & \textbf{64.36} & \textbf{0.1650} \\
            Ours & \checkmark & \checkmark & 69.09 & 37.02 & 21.34 & 63.02 & 0.1659 \\
    	\bottomrule
		\end{tabular}
    \vspace{-1.8em} 
}\end{table*}

Additionally, Fig. \ref{fig:viz_prob} presents qualitative results of contact probability heat-map obtained by VTLoc, showcasing several representative objects. For each object, we visualize the probability map and the ground truth contact location for reference.
The 3D heat-maps demonstrate a high degree of interpretability and reliability, offering an intuitive understanding of contact likelihood across object surface. Remarkably, VTLoc exhibits robust performance in challenging cases, such as intricate geometry of the swan, highlighting its strong generalization capability across diverse and complex shapes. These results emphasize the effectiveness of our method in achieving accurate and interpretable tactile contact localization. 

\subsection{Contact Localization with Symmetry Priors}
\begin{figure} [!t]
	\centering
    \vspace{0.4em}
	\includegraphics[width=0.6 \linewidth]{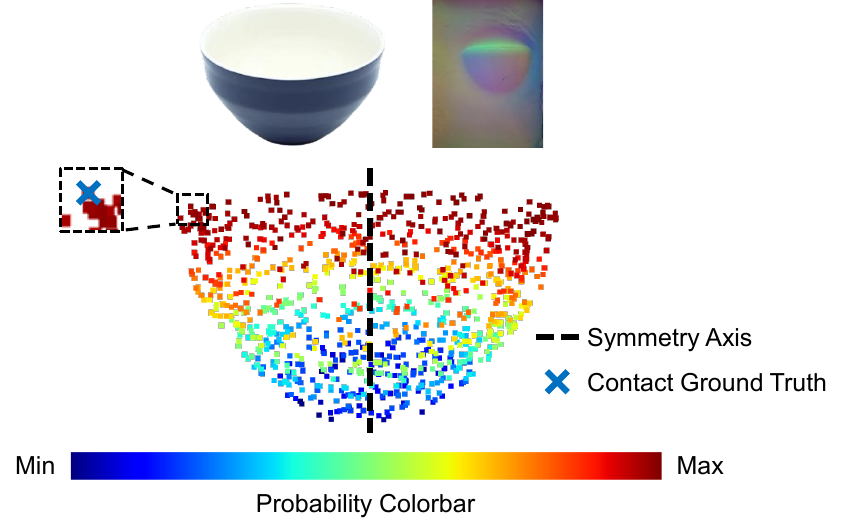}
    \vspace{-0.8em}
	\caption{
        Qualitative contact probability heatmap example of a bowl with rotational symmetry prior, demonstrating that our VTLoc can address the inherent symmetrical ambiguity of contact localization itself. \zhiyuanrevision{The symmetry axis is manually specified from the known object geometry.}
        }
    \label{fig:symmetry}
    \vspace{-0.8em}
\end{figure}

Many daily objects, like bowls and plates, exhibit geometric symmetries, creating inherent ambiguity in contact localization, as symmetric locations produce identical tactile patterns. Referring to previous works like \cite{bonzini2025symmetry}, our method can address this inherent ambiguity with symmetry priors. As shown in Fig. \ref{fig:symmetry}, by applying rotational mapping of the predicted contact candidates around the symmetry axis, we generate a rotationally symmetric contact probability heatmap that better reflects all plausible contact locations on the symmetric bowl.

\subsection{Multi-Contact Localization with VTLoc}

\begin{table*}[!t]
    \vspace{0.4em}
    \centering\footnotesize{
    \caption{
        Quantitative comparison of multi-contact localization results on the two subsets.
    }
    \vspace{-0.8em}
    \label{tab:multi_contact}
    \begin{tabular}
        {c c c c c c c}
        \toprule
        Subsets & Methods & mm Error (mm) $\downarrow$ & ND (\%) $\downarrow$ & Top-1 Acc (\%) $\uparrow$ & Top-5 Acc (\%) $\uparrow$ & MPE $\downarrow$ \\
        \hline
        \multicolumn{1}{l}{\multirow{2}{*}{Non-Uniform}} & MidasTouch & 16.22 & 8.76 & 72.96 & 99.37 & 0.0459 \\
        & \textbf{Ours} & \textbf{9.01}  & \textbf{4.85}  & \textbf{80.62} & \textbf{100.00} & \textbf{0.0323} \\
        \hdashline
        \multicolumn{1}{l}{\multirow{2}{*}{Uniform}} & MidasTouch & 35.17 & 18.84 & 50.95 & 85.81 & 0.0937 \\
        & \textbf{Ours} & \textbf{28.86} & \textbf{15.49} & \textbf{61.45} & \textbf{94.75} & \textbf{0.0718} \\
        \bottomrule
    \end{tabular}
    \vspace{-1.2em}
}\end{table*}

To investigate whether the single-contact improvement of VTLoc can be extended to multi-contact localization, we evaluate VTLoc with sequential tactile observations. Following the Monte Carlo filtering in MidasTouch  \cite{suresh2023ycbslide}, we replace the regression decoder with a probabilistic scoring head over the discrete contact candidates, producing a likelihood distribution for each tactile observation. For each target contact, we use the tactile images and relative motions from the three preceding contact indices in the data-collection order as sequential inputs for belief update. As shown in Tab.~\ref{tab:multi_contact}, incorporating multiple contacts improves localization performance compared with the single-contact setting. For example, on the non-uniform subset, the ND metric decreases from 20.17\% to 4.85\%. Furthermore, VTLoc consistently outperforms MidasTouch across all evaluation metrics on both subsets. On the non-uniform subset, VTLoc achieves an ND of 4.85\%, compared with 8.76\% for MidasTouch, representing an improvement of 3.91\%. The performance gains are more pronounced on the non-uniform subset, where richer local geometric variations provide additional cues that can be accumulated across sequential tactile observations. These results demonstrate that the stronger single-contact localization capability of VTLoc translates into improved multi-contact localization through more accurate likelihood estimation and belief updates.

\subsection{Ablation Study}  
We perform ablation studies to evaluate the effectiveness of our GMA and ILU modules. Objects with non-uniform surface curvature variations demonstrate more pronounced differences, making them better suited for assessing contact localization performance. Therefore, we utilize the `non-uniform' subset for our ablation studies.

\begin{table*}[!t]
	\centering\footnotesize{
	\caption{
        Ablation study on GMA with different modalities. We present statistical quantitative results comparing VTLoc with and without the GMA module, as well as evaluate performance across different combinations of input modalities. 
    }
    \vspace{-0.8em} 
\label{tab:gma_modalities}
	\begin{tabular}
            {l c c c c c c c c c}
            \toprule
            \multicolumn{1}{c}{\multirow{2}{*}{Methods}} & \multicolumn{2}{c}{Modality} & \multicolumn{1}{c}{\multirow{2}{*}{ND (\%) $\downarrow$}} & \multicolumn{1}{c}{\multirow{2}{*}{Top-1 Acc ($\%$) $\uparrow$}} & \multicolumn{1}{c}{\multirow{2}{*}{Top-5 Acc ($\%$) $\uparrow$}} & \multicolumn{1}{c}{\multirow{2}{*}{MPE $\downarrow$}} \\
            & Coordinates & Normals \\
            \hline
            \multicolumn{1}{l}{\multirow{2}{*}{w/o GMA}} & \checkmark &  & 22.58 $\pm$ 0.68 & 34.91 $\pm$ 1.47 & 82.36 $\pm$ 1.57 & 0.1232 $\pm$ 0.0057 \\
            & \checkmark & \checkmark & 22.29 $\pm$ 1.78 & 33.65 $\pm$ 3.22 & 81.13 $\pm$ 2.28 & 0.1227 $\pm$ 0.0094 \\
            \cdashline{1-7}
            \multicolumn{1}{l}{\multirow{2}{*}{w/ GMA}} & \checkmark &  & 21.69 $\pm$ 0.54 & 36.48 $\pm$ 1.60 & 83.28 $\pm$ 2.12 & 0.1171 $\pm$ 0.0101 \\
            & \checkmark & \checkmark & \textbf{20.50 $\pm$ 0.48} & \textbf{41.35 $\pm$ 2.90} & \textbf{85.22 $\pm$ 1.36} & \textbf{0.1091 $\pm$ 0.0054} \\
    	\bottomrule
		\end{tabular}
    \vspace{-1.0em} 
}\end{table*}

\subsubsection{Effectiveness of the GMA Module}  
\label{exp:gma_modalities}  
To assess the effectiveness of the GMA module, we perform an ablation study on different input modalities, \textit{i.e.}, coordinates and normal vectors. As shown in Tab. \ref{tab:gma_modalities}, VTLoc without GMA shows no improvement when adding normals to coordinates, and for some metrics, performance degrades (e.g., a drop in Top-1 and Top-5 Acc), likely due to the baselines lacking strong analytical capabilities for multi-modal point clouds. In contrast, VTLoc with GMA achieves improvements of 1.19\% in ND, 4.87\% in Top-1 Acc, 1.94\% in Top-5 Acc, and 6.8\% in MPE compared with using coordinates alone. Adding normals further enhances performance, as they provide richer geometric information. 

\subsubsection{Impact of Iteration Number in ILU}  
\label{sec:exp_gru}  
We conduct an ablation study to evaluate the effect of the iteration number $N$ in ILU. 
As shown in Tab.~\ref{tab:gru}, compared with the baseline ($N=0$), using $N=16$ reduces ND from 22.97\% to 20.17\%, increases Top-1 Acc from 29.56\% to 44.65\% and Top-5 Acc from 79.87\% to 86.16\%, and reduces MPE from 0.1129 to 0.1037.
Performance improves consistently as $N$ increases, peaking at $N = 16$, but declines beyond this point due to over-iteration, which leads to overfitting and reduces generalization. 

\begin{table*}[!t]
    \vspace{0.4em}
	\centering\footnotesize{
	\caption{
        Ablation study on ILU iteration numbers. We present both evaluation metrics and inference time to provide a comprehensive assessment. 
    }
    \vspace{-0.8em} 
\label{tab:gru}
	\begin{tabular}
            {c c c c c c}
            \toprule
            Iteration Numbers & ND (\%) $\downarrow$ & Top-1 Acc ($\%$) $\uparrow$ & Top-5 Acc ($\%$) $\uparrow$ & MPE $\downarrow$ & Inference Time (ms) $\downarrow$ \\
            \hline
            0 & 22.97 & 29.56 & 79.87 & 0.1129 & \textbf{123.56} \\
            4 & 20.80 & 37.74 & 84.28 & 0.1101 & 123.92 \\
            8 & 20.72 & 38.99 & 85.53 & 0.1049 & 125.31 \\
            16 & \textbf{20.17} & \textbf{44.65} & \textbf{86.16} & \textbf{0.1037} & 128.05 \\
            32 & 20.56 & 40.88 & 85.53 & 0.1214 & 134.29 \\
            64 & 21.61 & 36.48 & 83.65 & 0.1279 & 145.99 \\
    	\bottomrule
		\end{tabular}
    \vspace{-1.8em} 
}\end{table*}

It is noteworthy that VTLoc maintains a total parameter size of 124.28 MiB and an average inference time of 128.05 ms. Increasing $N$ has no impact on the parameter count and adds only 4.49 ms to inference time compared to the baseline ($N=0$). These results demonstrate that our approach is efficient and practical for real-world robotics applications.  

\subsubsection{Contact Localization for Unknown Objects}
We also conduct experiments on unknown objects: splitting the data into training and testing sets by the object level in a 7:3 ratio. As reported in Tab. \ref{tab:unknown}, our method shows superior performance against baselines, achieving significant improvements in both Top-1 and Top-5 accuracy metrics while maintaining the lowest MPE score among all methods.
\begin{table}[!t]  
	\centering\footnotesize{
	\caption{
        Quantitative results of our proposed VTLoc for unknown objects. 
    } \label{tab:unknown}
    \vspace{-0.8em} 
    \begin{tabular}{l|c|c|c|c}  
    \toprule  
        Methods & ND (\%) $\downarrow$ & Top-1 Acc $\uparrow$ & Top-5 Acc $\uparrow$ & MPE $\downarrow$ \\
        \hline  
        Chance & 98.95 & 2.88 & 16.04 & 0.2727 \\
        Point Filtering & 44.47 & 3.26 & 15.22 & 0.2594 \\
        MCR & 42.27 & 4.32 & 16.39 & 0.2403 \\
        \cdashline{1-5}  
        \textbf{Ours} & \textbf{36.69} & \textbf{8.65} & \textbf{32.79} & \textbf{0.1883} \\
    \bottomrule  
    \end{tabular}  
    \vspace{-1.8em} 
}\end{table}  

\section{Conclusion and Discussion} 
\label{sec:conclusion}
In this paper, we present VTLoc, a novel framework for visual-tactile contact localization in point cloud that estimates contact localization probability of each point by matching tactile features with the geometric features of the point cloud. We introduce a geometric multi-modal alignment (GMA) module to align tactile and visual modalities by leveraging geometric information and an iterative localizing updater (ILU) that performs contact location regression in an iterative way. To validate our methods, we establish a visual-tactile point cloud contact localization benchmark comprising 100 real-world objects. Our quantitative and qualitative results show that VTLoc improves single-touch contact localization by reducing local-to-global correspondence ambiguity through geometric multi-modal alignment and iterative localization refinement. 

Despite the great effectiveness demonstrated by our VTLoc, it still has limitations. Contact localization is inherently ambiguous, as similar tactile readings can arise from different contact locations, \textit{e.g.}, touching a flat surface at the same pose. In such cases, performing localization iteratively across multiple contacts can help reduce uncertainty and improve downstream tasks such as object pose estimation and manipulation.
Additionally, we only \zhiyuanrevision{focus} on hard objects. When extending to soft objects, temporal tactile sequences can be considered to capture deformation dynamics during contact.

{
\tiny
\bibliographystyle{IEEEtran}
\bibliography{references}
}

\end{document}